\documentclass{ieeeaccess}
\usepackage{cite}
\usepackage{amsmath,amssymb,amsfonts}
\usepackage{graphicx}
\usepackage{textcomp}

\usepackage{algorithmic}
\usepackage{algorithm}
\usepackage{array}
\usepackage[caption=false,font=normalsize,labelfont=sf,textfont=sf]{subfig}
\usepackage{stfloats}
\usepackage[hyphens]{url}
\usepackage{verbatim}
\usepackage{acronym}
\usepackage{multirow, multicol}
\usepackage{mathabx}
\usepackage{makecell}
\usepackage{caption}
\usepackage{soul}
\usepackage{csquotes}
\usepackage{mdframed}

\newacro{kpi}[KPI]{Key Performance Indicator}
\newacro{tdoc}[Tdoc]{Technical Documents}
\newacro{cr}[CR]{Change Request}
\newacro{rf}[RF]{Radio Frequency}
\newacro{ran}[RAN]{Radio Access Network}
\newacro{sa}[SA]{system architecture}
\newacro{ct}[CT]{core network and terminals}
\newacro{qos}[QoS]{quality of service}
\newacro{wg}[WG]{working group}
\newacro{5ga}[5G-A]{5G advanced}
\newacro{bpe}[BPE]{byte pair encoding}
\newacro{mmb}[MMB]{multi-modal beamforming}
\newacro{jscc}[JSCC]{joint source-channel coding}
\newacro{sft}[SFT]{Supervised Fine-Tuning}
\newacro{dpo}[DPO]{Direct Preference Optimization}
\newacro{llm}[LLM]{Large Language Model}
\newacro{lmm}[LMM]{Large Multi-modal Model}
\newacro{fm}[FM]{Foundation Model}
\newacro{ai}[AI]{Artificial Intelligence}
\newacro{lm}[LM]{Language Modeling}
\newacro{ptlm}[PTLM]{Pre-Trained Language Model}
\newacro{nlp}[NLP]{Natural Language Processing}
\newacro{dl}[DL]{Deep Learning}
\newacro{nn}[NN]{Neural Network}
\newacro{dnn}[DNN]{Deep Neural Network}
\newacro{cnn}[CNN]{Convolutional Neural Network}
\newacro{rnn}[RNN]{Recurrent Neural Network}
\newacro{gnn}[GNN]{Graph Neural Network}
\newacro{ml}[ML]{Machine Learning}
\newacro{cv}[CV]{Computer Vision}
\newacro{ssl}[SSL]{Self-Supervised Learning}
\newacro{tl}[TL]{Transfer Learning}
\newacro{nlm}[NLM]{Neural Language Model}
\newacro{lstm}[LSTM]{Long Short-Term Memory}
\newacro{gpt}[GPT]{Generative Pre-trained Transformer}
\newacro{bert}[BERT]{Bidirectional Encoder Representation from Transformer}
\newacro{nlu}[NLU]{Natural Language Understanding}
\newacro{nlg}[NLG]{Natural Language Generation}
\newacro{t5}[T5]{Text-to-Text Transfer Transformer}
\newacro{icl}[ICL]{In-Context Learning}
\newacro{rlhf}[RLHF]{Reinforcement Learning with Human Feedback}
\newacro{mha}[MHA]{Multi-Head Attention}
\newacro{clm}[CLM]{Causal Language Modeling}
\newacro{mlm}[MLM]{Masked Language Modeling}
\newacro{plm}[PLM]{Permuted Language Modeling}
\newacro{dae}[DAE]{Denoising AutoEncoder}
\newacro{rf}[RF]{Radio Frequency}
\newacro{sota}[SOTA]{state of the art}
\newacro{rag}[RAG]{Retrieval Augmented Generation}
\newacro{moe}[MoE]{Mixture of Expert}
\newacro{peft}[PEFT]{Parameter-Efficient Fine-Tuning}
\newacro{sdo}[SDO]{Standards Developing Organization}
\newacro{cot}[CoT]{Chain-of-Thought}
\newacro{rl}[RL]{Reinforcement Learning}
\newacro{vlm}[VLM]{Visual Language Model}
\newacro{6g}[6G]{Sixth Generation}
\newacro{cv2x}[CV2X]{Cellular Vehicle-to-Everything}
\newacro{esti}[ESTI]{European Telecommunication Standards Institute}
\newacro{oran}[O-RAN]{Open Radio Access Network}
\newacro{qos}[QoS]{Quality of Service}
\newacro{3gpp}[3GPP]{Third Generation Partnership Project}
\newacro{itu}[ITU]{International Telecommunication Union}
\newacro{ran}[RAN]{Radio Access Network}
\newacro{bs}[BS]{Base Station}
\newacro{its}[ITS]{Intelligent Transport System}
\newacro{rrm}[RRM]{Radio Resource Management}
\newacro{lora}[LoRA]{Low Rank Adaptation}
\newacro{mlp}[MLP]{Multi-Layer Perceptron}
\newacro{vit}[ViT]{Vision Transformer}
\newacro{qat}[QAT]{Quantization Aware Training}
\newacro{ptq}[PTQ]{Post-Training Quantization}
\newacro{kv}[KV]{Key-Value}
\newacro{rleif}[RLEIF]{Reinforcement Learning from Evol-Instruct Feedback}
\newacro{v2x}[V2X]{Vehicle to Everything}
\newacro{rag}[RAG]{Retrieval Augmented Generation}
\newacro{fim}[FIM]{Fill-In-the-Middle}
\newacro{mcq}[MCQ]{Multiple-Choice Question}
\newacro{qa}[QA]{Question Answering}
\newacro{ieee}[IEEE]{Institute of Electrical and Electronics Engineers}
\newacro{urllc}[URLLC]{Ultra Reliable and Low Latency Communication}
\newacro{kl}[KL]{Kullback-Leibler}
\newacro{cdf}[CDF]{Cumulative Density Function}
\newacro{rrm}[RRM]{Radio Resource Management}
\newacro{genai}[GenAI]{Generative Artificial Intelligence}
\newacro{agi}[AGI]{Artificial General Intelligence}
\newacro{kg}[KG]{Knowledge Graph}
\newacro{kb}[KB]{Knowledge Base}
\newacro{ve}[VE]{Vector Embedding}
\newacro{te}[TE]{Topological Embedding}
\newacro{bs}[BS]{Base Station}
\newacro{ue}[UE]{User Equipment}
\newacro{av}[AV]{Autonomous Vehicle}
\newacro{ci}[CI]{Collective Intelligence}
\newacro{rl}[RL]{Reinforcement Learning}

\def\BibTeX{{\rm B\kern-.05em{\sc i\kern-.025em b}\kern-.08em
    T\kern-.1667em\lower.7ex\hbox{E}\kern-.125emX}}

\begin{document}
\history{Date of publication April 30, 2025, date of current version April 30, 2025.}
\doi{10.1109/ACCESS.2025.3565859}

\title{GenAINet: Enabling Wireless Collective Intelligence via Knowledge Transfer and Reasoning}
\author{
\uppercase{Hang Zou}\authorrefmark{1},
\uppercase{Qiyang Zhao} \authorrefmark{1}, 
\uppercase{Samson Lasaulce} \authorrefmark{2,4}, \IEEEmembership{Member, IEEE},
\uppercase{Lina Bariah} \authorrefmark{2}, \IEEEmembership{Senior Member, IEEE}, 
\uppercase{Mehdi Bennis} \authorrefmark{3}, \IEEEmembership{Fellow, IEEE},
\uppercase{and M{\'e}rouane Debbah} \authorrefmark{2}, \IEEEmembership{Fellow, IEEE} 
}
\address[1]{Artificial Intelligence and Digital Science Research Center, Technology Innovation Institute, 9639 Abu Dhabi, UAE}
\address[2]{6G Research Center, Khalifa University, 127788 Abu Dhabi, UAE }
\address[3]{Centre for Wireless Communications, University of Oulu, 90570 Oulu, Finland}
\address[4]{CRAN, Université de Lorraine and CNRS, 54506 Vandœuvre-lès-Nancy, France}


\markboth
{H. Zou \headeretal: GenAINet: Enabling Wireless Collective Intelligence via Knowledge Transfer and Reasoning}
{H. Zou \headeretal: GenAINet: Enabling Wireless Collective Intelligence via Knowledge Transfer and Reasoning}

\corresp{Corresponding author: Samson Lasaulce (samson.lasaulce@univ-lorraine.fr)}

\begin{abstract}

\Ac{genai} and communication networks are expected to have groundbreaking synergies for 6G. Connecting \ac{genai} agents via a wireless network can potentially unleash the power of \acf{ci} and pave the way for \acf{agi}. However, current wireless networks are designed as a ``data pipe'' and are not suited to accommodate and leverage the power of \ac{genai}. In this paper, we propose the GenAINet framework in which distributed \ac{genai} agents communicate knowledge (facts, experiences, and methods) to accomplish arbitrary tasks. We first propose an architecture for a single \ac{genai} agent and then provide a network architecture integrating GenAI capabilities to manage both network protocols and applications. Building on this, we investigate effective communication and reasoning problems by proposing a semantic-native GenAINet. Specifically, GenAI agents extract semantics from \textcolor{black}{heterogeneous} raw data, build and maintain a knowledge model representing the semantic relationships among pieces of knowledge, which is retrieved by GenAI models for planning and reasoning. 
\textcolor{black}{Under this paradigm, different levels of collaboration can be achieved flexibly depending on the complexity of targeted tasks.}
Furthermore, we conduct two case studies in which, through wireless device queries, we demonstrate that extracting, \textcolor{black}{compressing} and transferring common knowledge can improve query accuracy while reducing communication costs; and in the wireless power control problem, we show that distributed agents can \textcolor{black}{complete general tasks independently through collaborative reasoning} \textcolor{black}{without predefined communication protocols}. 
Finally, we discuss challenges and future research directions in applying \acp{llm} in 6G networks.
\end{abstract}


\titlepgskip=-15pt

\maketitle

\section{Introduction}
\label{sec:introduction}
\PARstart{T}{he} concept of \Acf{ci} has been discussed for many years, including within the wireless community for instance, \textcolor{black}{as explored in works such as \cite{narayanan2022access,li2020twc}.} \ac{ci} is a form of decision-making where intelligent, distributed human and software agents, situated in a networked communication system, receive information and feedback from their immediate environment and other agents and make decisions collectively to perform tasks that, together, achieve a common desirable outcome \cite{angelo2019seams}. However, the architecture of related works either remains at the conceptual level in \cite{narayanan2022access} or with pre-defined meaning to exchange as in \cite{li2020twc}. Existing frameworks severely limit the applications of \ac{ci}. First, tasks that can accomplished in these frameworks are both {limited and fixed}. If new tasks arise due to the environment change or new demands from \acfp{ue}, it would be extremely costly and difficult to adapt existing system to a new scenario. Second, {data heterogeneity} prevents a unified structure of the individual agent from being implemented. Third, the interaction mechanism between agents is pre-defined, limiting {flexibility}. Finally the information exchange between agents occurs only in a {raw data space}, leading to useless and possibly huge communication costs e.g., regarding the final task for the network or communication. Therefore, a new paradigm of \ac{ci} is required to alleviate these drawbacks. 

Meanwhile, 6G wireless networks are envisioned to be AI-native, in the sense that wireless communications will be an integrated part of training and inference. Conventionally, wireless networks are designed for data collection and transmission, aiming at achieving a targeted \acf{qos}. However, they are not designed to support a massive deployment of AI-empowered devices, especially with large AI models which have high communication and computing costs, and where a wide range of tasks have to be executed. \Acfp{llm} built on \acfp{gpt} have shown impressive capabilities, from question answering and language understanding, to mathematical and common sense reasoning \cite{yenduri2023generative}. Such capabilities facilitate the wide adoption of \acp{llm} in Telecom \cite{bariah2023understanding, Lina2024,Xu2024LargeMM}, finance \cite{liu2023fingpt,wu2023bloomberggpt}, healthcare \cite{singhal2023towards}, and so on. The maturity of \acp{llm}, which is achieved by training on massive data and compute, constitutes an important step towards \acf{agi}. 
Connecting distributed \acp{llm} through wireless networks paves the way to enable multi-agent  \ac{ci} \cite{Lina2024}. There are three main advantages possessed by \acp{llm} compared to conventional \acfp{dnn}: emerging abilities \cite{wei2022emergent,brown2020gpt3,wei2021finetuned}, reasoning $\&$ planning \cite{wei2022chain}, and autonomous cooperation \cite{li2023camel,park2023generative} as a potential enabler for \ac{ci} in 6G era.

\textbf{Emerging Abilities} refers to the capabilities of \acp{llm} that appear suddenly and unpredictably as model size, computational power, and training data scale up. There are two typical emergent capabilities \acp{llm} have been to able to perform: in-context learning \cite{brown2020gpt3} and instruct following \cite{sanh2022multitask,ouyang2022training,wei2021finetuned}. In context learning is that \acp{llm} are capable to perform unseen tasks provided with several demonstrative examples or the the description of the task in natural language. Such capability enable agents to accomplish different tasks in a varying environment without re-training or fine-tuning. Besides, \acp{llm} can follow task instructions described in natural language without using explicit examples. These capabilities demonstrate the strong generalization ability of \acp{llm}, making them suitable for diverse tasks in wireless networks.

\textbf{Reasoning and Planning} \acp{llm} can perform logical or math reasoning to a certain level as showed in \cite{luo2023wizardmath,shao2024deepseekmath}. For instance, \acp{llm} can leverage \acfp{cot} prompting, i.e., to obtain a final answer via multi-step intermediate reasoning state \cite{wei2022chain}.  The capability of reasoning and planning are crucial to enable collective intelligence because: i) tasks in telecom context, e.g., \acf{rrm} generally requires logical and math reasoning; ii) it is essential to decompose a high-level goal/intent to known executable tasks and plan them in a reasonable sequential order.

\textbf{Autonomous Cooperation} \acp{llm} can be further empowered into autonomous agents with external memory and tool utilization. An agent can observe the environment, plan a sequence of actions, create high-level reflections of experience in a memory stream, and formulate plans. For example, BabyAGI is a task-driven autonomous agent framework that can generate, execute and prioritize tasks in real-time \cite{babyagi}.  Similarly, Auto-GPT can chain together \ac{llm} ``thoughts'' in an infinite loop of reasoning \cite{autogpt}. For solving even more complex tasks with \acp{llm}, which are beyond the capability of a single agent, a multi-agent framework is required.   On the one hand, multi-agent systems are shown to largely enhance \acp{llm}' capability in task solving \cite{li2024agents}. For example, CAMEL explores two role-playing communicative agents with autonomous cooperation \cite{li2023camel}. In generative agents, believable human behaviors are simulated in a sandbox environment, where multiple agents interact in natural language to complete spontaneous tasks \cite{park2023generative}. On the other hand, adversarial interactions can also enhance overall performance through multi-agent debate as shown in \cite{liang2023encouraging,du2023improving}. However, communication, computing, and storage efficiency are not considered in these frameworks, which are critical factors in wireless networks.


Motivated by the advantages bringing by \acp{llm} above and the drawbacks of the plug-in deployment of \acp{llm} in a wireless network, in this paper, we propose the GenAINet \footnote{We adopt the terminology \ac{genai} to refer to all generative foundation models including \acp{llm} to maintain flexibility.} framework. In our framework, agents leverage multi-modal \acp{llm} to extract semantics in common embedding space from different data modalities. Then multiple GenAI agents exchange knowledge to perform effective reasoning and solve arbitrary tasks of particular applications, e.g., networks of vehicles, smart grid, and internet of things through effective semantic communication. 
The agents communicate knowledge to help each other's planning and decision as well. In doing so, the performance can be improved with less communication and computing costs. The main contributions of this article are as follows:

\begin{itemize} 

\item \textcolor{black}{We propose a novel GenAINet architecture that overcomes the drawback of existing frameworks by enabling dynamic task adaptation. Unlike current systems that are limited to fixed tasks, GenAINet integrates \acp{llm} to handle evolving and arbitrary tasks arising from environmental changes or new user demands. This flexibility ensures the system remains adaptive and cost-efficient in dynamic scenarios.} 

\item \textcolor{black}{We address the challenge of data heterogeneity by proposing a unified agent architecture capable of managing both network protocols and applications. This unified design allows for seamless integration of knowledge and decision-making mechanisms across diverse agents, eliminating the structural limitations of existing frameworks. Interaction between agents is no longer pre-defined; instead, reasoning and communication occur dynamically based on task requirements.} 

\item \textcolor{black}{A communication paradigm for GenAINet is introduced, built on environmental semantics and multi-path reasoning. GenAINet mitigates the inefficiencies of raw data exchange by employing semantic-level communication, where semantics are extracted uniformly from multi-modal data using multi-modal \acp{llm}. This reduces communication costs while retaining task-relevant information. Through multi-step planning, GenAINet leverages \acp{llm} to optimize reasoning paths toward effective decisions, allowing experiences and mechanisms to be shared across agents to improve independent or collaborative planning.} 

\item \textcolor{black}{Two case studies demonstrate GenAINet’s versatility in addressing limitations of traditional systems. In the first use case, semantic remote query on a mobile device, we show that transferring common knowledge from a teacher to a student agent improves query accuracy and reduces communication overhead through LLMs' semantic compression. In the second use case, distributed wireless power control, we show that GenAI agents can solve power allocation problems independently, for arbitrary performance metrics, through collaborative reasoning, outperforming traditional fixed-policy approaches.} 

\item \textcolor{black}{We explore challenges and future research directions, including deploying \acp{llm} on edge devices, building hierarchical world models, and optimizing agent interactions to balance flexibility, scalability, and efficiency. These advancements position GenAINet as a transformative paradigm for addressing the limitations of existing wireless networks and AI approaches.}  

\end{itemize}

\section{GenAI Network and Agent Architectures}

\begin{figure*}[t!]
\centering
\includegraphics[width=1.0\linewidth]{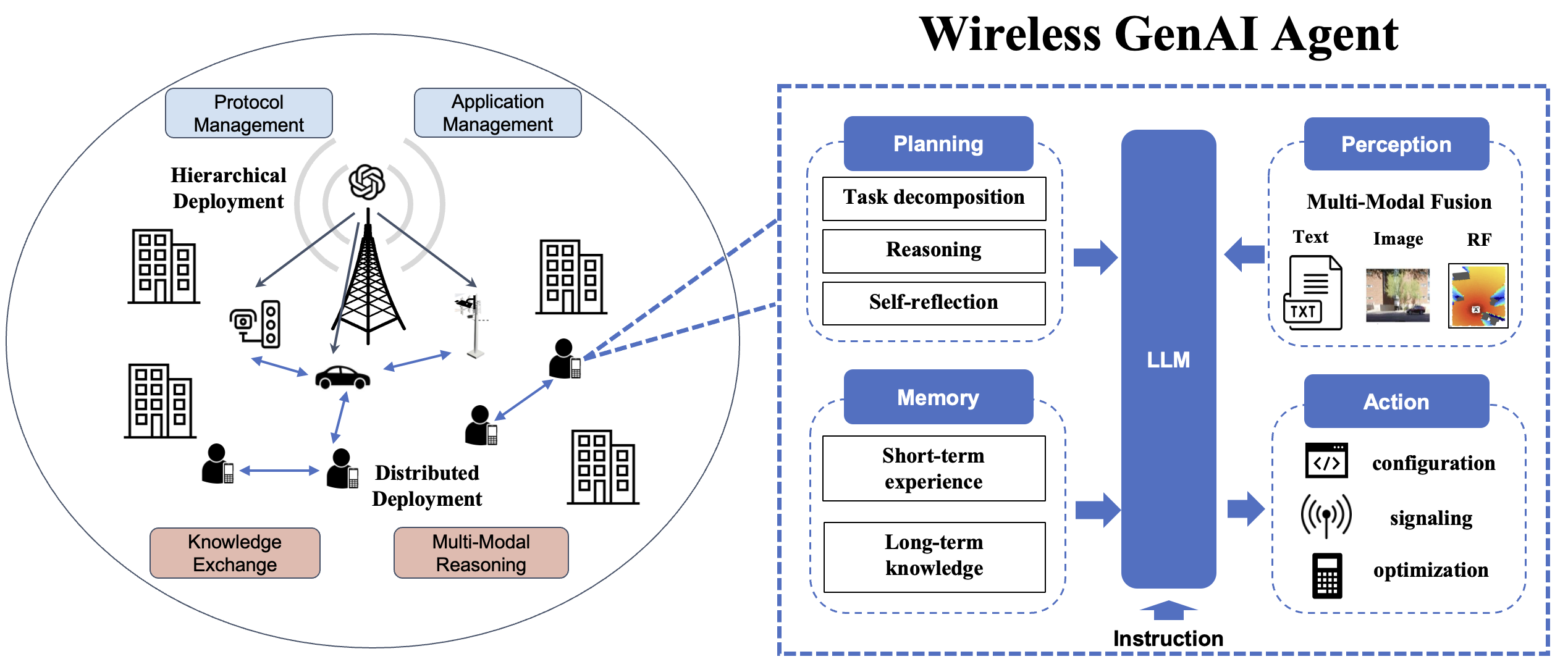}
\caption{
\textcolor{black}{Proposed GenAINet architecture (left) and agent architecture (right) for protocol and application management. GenAINet leverages unified GenAI agents to manage communication protocols (e.g., PHY, MAC) and applications (e.g., robotics, vehicles). Each device is associated with an agent that communicates using semantic knowledge. The GenAI agent includes four components: perception (collects information from the environment and other agents), memory (stores long/short-term knowledge in formats like vector embeddings and knowledge graphs), action (available tools), and planning (reasoning and decision-making using LLMs).}}
\label{fig:GenAINetArchitecture}
\end{figure*}

Wireless \ac{genai} agents can emulate human-like decision making process, providing a path towards sophisticated and adaptive GenAI networks. It can bring autonomy to network protocols and network applications. 
d in Fig. \ref{fig:GenAINetArchitecture}.
To achieve this, we propose prospect architectures for GenAI agents, as shown in Fig. \ref{fig:GenAINetArchitecture}.

\subsection{Wireless GenAI Agent Architecture}
\label{sec:agent_structure}
To enable LLMs to interact with real-world scenarios, an agent has to be of a distributed structure, decision-wise and information-wise. An autonomous agent generally comprises four components: perception, action, planning, and memory \cite{wang2024survey,xi2023rise}. In what follows, we explain how these components can be designed to implement the proposed vision and discuss their roles in wireless networks. 

\textbf{Perception} is the part of an agent gathering useful and relevant information to an \ac{llm}. Possible information sources for perception include: 1) information sensed from the environment, e.g., channel state information (CSI), locations of \ac{ue} and \ac{bs} in a network, or on an autonomous car, the traffic density, speed and visual data of streets; 2) information received from other agents, such as messages reporting their state or responses to some requested information, or strategies learned from past decisions under different states. The information can be represented by using different modalities, from raw text, images, to abstracted graphs. The perceptor should fuse and encode raw data in a common embedding space, for the LLM to perform further decision-making.

\textbf{Action} is the component managing all available tools of an agent. Actions generated by the LLM should be adapted to the target functions or interfaces to execute. For a wireless \ac{genai} agent, possible actions may include: 1) information generated to complete a task, such as the responses to a query from user, or messages sent to another agent to complete a task; 2) executing predefined interfaces and toolkits, such as the function call in a MAC scheduler, configurations in the radio amplifier, or commands to adjust the car speed and steering. Furthermore, the action in a multi-agent network includes actions taken by observing the environment and communication actions (messages) to be sent to other agents. Both should be generated from a learned joint strategy. 

\textbf{Planning} is the process of creating and optimizing actions over time to achieve a goal including: 1) sub-task decomposition, i.e.,  breaking down a high-level goal into sub-goals and actionable tasks; 2) synchronize and prioritize existing tasks between agents; 3) self-reflection, which evaluates and criticizes the past decisions to optimize the policy. Planning allows \acp{llm} to solve complex tasks via multi-step reasoning, such as \acp{cot}. Furthermore, such planning and reasoning can be conducted collaboratively among multiple agents by leveraging computing resources and tools on different devices efficiently.  

\textbf{Memory} is the module for storing short-term experience (evolving knowledge)  and long-term knowledge (common knowledge) for \acp{llm}' future planning and decisions. 
Short-term experience is the selected history of observations, actions, thoughts, and conclusions of agents. It is necessary for agents to efficiently reuse or adapt existing solutions or mechanisms instead of reasoning and extracting from scratch.
Moreover, the long-term knowledge contains facts, methods, and contents which could be general or domain-specific depending on the role of an agent. Having such knowledge could effectively reduce the chance of an LLM to produce non-reliable or non-factual responses by using retrieval-augmented techniques. Possible knowledge representations and how \acp{llm} leverage them to enhance the quality of inference will be covered in next section. Empowered by \acp{llm}' knowledge, \ac{genai} agents can release the need for environment modeling or problem formulation, while finding the reasoning path towards task completion. This makes \ac{genai} agents effective in handling complex, unseen problems.

\subsection{Wireless GenAI Network Architecture}
In a wireless network, a \ac{genai} agent can potentially act as an autonomous controller in the following scenarios.

\textbf{Network protocols}: \ac{genai} agents can orchestrate network resources and control network functions in radio access, core, and transport domains. It can interface with network protocols to control power, resource, traffic, etc. The GenAI network protocol architecture can include: 
\begin{itemize}
    \item \textit{Hierarchical architecture}: \ac{genai} agents can be deployed in multi-level network controllers, sending policies to the data network. For instance, a commander agent operates in the RAN intelligent controller (RIC), while multiple executor agents operate in the network elements (NE). The commander deconstruct intents into subtasks and assigns to the executors.
    \item \textit{Distributed architecture}: \ac{genai} agents can be co-located with distributed NEs and manage locally the data and control planes. The interaction between agents is self-organized, where agents exchange past decisions or knowledge to collaboratively accomplish the tasks. 
\end{itemize}

In autonomous networks, \ac{genai} agents can break down high-level intents, plan actionable tasks, generate network control strategies, and refine them from feedback. 

\textbf{Network applications}: \ac{genai} agents can bring autonomy to network applications, such as autonomous vehicles and robots. Wireless networks allow \ac{genai} agents to collaboratively perform various tasks, such as remote sensing, control, and planning. The GenAI network application architecture can include:
\begin{itemize}
    \item \textit{Independent architecture}: Each application is controlled by individual \ac{genai} agents. For example, in a vehicular network, the network agents control \acp{bs}, \acfp{ue} and the traffic agents control cars, traffic lights. 
    \item \textit{Converged architecture}: \ac{genai} agents can jointly control network protocols and applications. For example, a car agent can jointly control engine, steering and communication protocols with other cars, such that multiple cars can collaboratively improve the traffic flow. 
\end{itemize}

The hierarchical and independent architecture can be built over existing 5G wireless networks where applications are over-the-top services. On the other hand, the distributed and converged architectures are more disruptive for 6G, fully implementing the protocol and application convergence with multi-agent network. 

\textcolor{black}{With GenAI agents introduced, the network management plane will be enhanced with automated policy generation and adaptation powered by LLM reasoning. The control plane will be enhanced with automated cooperation protocols between agents.}

\section{Wireless GenAINet with Knowledge driven Communication and Reasoning}


Conventional communication systems operate on raw data. Moreover, agents memorize and exchange information from raw data regardless the relevance to their tasks. Besides, knowledge exchange between AI models on the network edge is typically done by exchanging model weights, gradients, or hyper-parameters, e.g., federated learning. However, such design schemes become inefficient when massive \ac{genai} agents with gigantic model size are connected. First, raw data contains a large amount of redundant information which is inefficient for communication; Second, adapting \acp{llm} over networks for different tasks is energy-consuming and may induce large latency. To address the aforementioned issues, we propose a semantic-native GenAINet. 
Pre-trained on massive amount of data, \acp{llm} can understand the context of a problem. Therefore, it would be sufficient to merely exchange knowledge and information in brief/detailed textual description between \ac{genai} agents to accomplish a goal collaboratively. Furthermore, wireless GenAINet being constrained by communication, computing, and storage resources, knowledge transferred and stored in the network should satisfy minimality and sufficiency, that a minimal amount of knowledge and information can solve a wide range of tasks or problems effectively. That said, how to efficiently exchange and store knowledge and information would be the key problem to enable wireless \ac{ci}.

\subsection{Efficient Knowledge Representation}
\acp{llm} have been trained to understand semantics, the meaning of a message. This allows \acp{llm} to effectively compose information for different tasks, leading to better generalization. {Knowledge can be categorized into two main types: a) common knowledge in a domain including facts, known methods to solving problems, predefined standards and so on; b) evolving knowledge that builds on top of common knowledge for specific tasks, including useful experiences and effective mechanisms that \acp{llm} have learned during operation. This highly abstracted knowledge should be efficiently shared among \acp{llm} in a network.} 
For example, common knowledge include predefined network parameter such as frequencies, bandwidth, and functions defined in \acf{3gpp} protocols and standards while evolving knowledge can be adaptive techniques for optimizing network performance in real-time scenarios, e.g., through \acf{rl}. 
In what follows, three possible representations for knowledge which are relevant for communication between \acp{llm} are discussed. 

\textbf{\Ac{ve}} is a commonly used knowledge representation for \acp{llm}. It embeds raw data (text, image, audio) with latent vectors to construct a vector database (DB), where the distance between vectors represents their semantic similarities. When an \ac{llm} receives a user prompt, it first leverages time-efficient methods,  e.g., approximate nearest neighbors, similarity search to locate relevant clusters based on similarity measure. A precise local research will be applied  to extract related information, which is concatenated with the user prompt for \acp{llm} to generate responses. \Acfp{rag} provide additional context to improve \acp{llm}'s performance in a time varying domain. \textcolor{black}{An example of VE in GenAINet can involve operational logs and deployment knowledge in a smart factory. Each log entry, such as equipment status, fault occurrences, or maintenance actions, is embedded into a vector space, where semantically similar logs (e.g., similar fault patterns) are close in distance. When an LLM agent receives a query about recent machine failures, it uses VE techniques like approximate nearest neighbors to quickly identify related logs, retrieves detailed maintenance actions from the corresponding clusters, and combines this information with the query to generate a contextual response. This enables efficient troubleshooting in a dynamic, time-varying operational environment.}

\textbf{\Ac{kg}} is a structured representation of relations between real objects or abstracted concepts. These relations can describe {connections} and {causalities} between entities in natural language, making it easily accessible for an \ac{llm}. Two vertices would be connected  by an edge if they are correlated and the weight of the edge may indicate the degree of correlation. \ac{kg} can assist an \ac{llm} to perform fast information retrieval, and efficient reasoning based on the structure of entities. Moreover, \ac{kg} can also help \acp{llm} to understand and generate more contextually relevant and coherent responses for complex tasks requiring factual responses. \textcolor{black}{To construct and maintain a KG tailored for 3GPP/IEEE standards and technical papers, key entities such as technical terms, specifications, and network components are extracted using NLP techniques like named entity recognition (NER). Relationships between these entities, such as dependencies or causal links, are identified through pattern matching and contextual analysis. The KG is then built by representing entities as vertices and relationships as weighted edges, with weights reflecting their relevance or significance. Continuous updates are essential as new standards and technical papers are published, ensuring the KG stays relevant and accurate. This dynamic KG enables GenAINet to perform efficient reasoning and information retrieval, supporting tasks like optimizing 5G network slicing or addressing emerging challenges in 6G networks.}

\textbf{\Ac{te}}: Topological models, e.g., hypergraphs, simplicial complexes and cell complexes, can represent the intrinsic structure of data. They can model the low-order (vertices) and high-order (simplices, cells) relations between syntactic objects or abstracted concepts. Unlike conventional neural networks learning data structure on 1D sequence or 2D grid, \ac{te} can exploit more flexible and dynamic structure. Moreover, the high-order topological structure can represent more complex and causal relations, such as different word combinations by semantic features. Therefore, \ac{te} can exploit more implicit latent structures than \ac{kg}. The geometric features allows for flexible structure composition according to the changes of semantic features, which saves communication and computing resources. Nevertheless, due to the complexity of topological models, the exploration of such structure is still on early stage. \textcolor{black}{An example of applying TE in GenAINet can be seen in modeling dynamic network interactions in a 6G environment. For instance, a simplicial complex can represent communication relations among devices in a vehicular network, where vertices are individual vehicles, edges represent direct communication links, and higher-order simplices capture multi-device group interactions (e.g., cooperative sensing). By dynamically updating the simplicial complex based on semantic features like traffic patterns or interference levels, TE can efficiently model and adapt to changes in the network structure, enabling resource-efficient communication and computation while preserving critical causal relationships.}

\textcolor{black}{Vector embeddings, knowledge graphs (KG), and topological embeddings each offer distinct trade-offs in computational and communication complexity for knowledge representation. VE is computationally efficient, with low communication costs  due to fixed-size embeddings, making it ideal for lightweight, real-time applications. However, it may lose granularity during compression. KG provides interpretable relationships and structured reasoning, with moderate computational complexity but higher communication costs for transmitting subgraphs. In contrast, TE captures rich, high-order interactions through complex topological structures, but this comes at the expense of very high computational and communication overhead, making it suitable only for specialized tasks. VE is most effective for efficiency-focused applications, KG excels in interpretability and reasoning, and TE is valuable for modeling intricate relational data in highly specialized scenarios.}

\textcolor{black}{To ensure effective integration, a robust mechanism for real-time updates and dynamic adaptation of the knowledge base (KB) is essential. Real-time updates can be achieved by enabling feedback loops between the querying device and the cloud-based KB. When new or ambiguous queries are encountered, the device can send feedback to the cloud, prompting the cloud LLM to refine or expand the KB with additional information. This dynamic process ensures that the KB evolves to accommodate emerging queries and remains aligned with the latest telecom standards and user needs. To balance data compression and query accuracy, the KB construction involves compressing semantic information into low-dimensional embeddings that capture essential knowledge while minimizing communication overhead. As we will show later in our first case study, the data compression rate of queries in different categories varies from one to others. Before compression, such knowledge learned during service could be provided to cloud LLM to adjust the compression rate carefully. This leads to a significant reduction of communication overhead especially for queries which are less sensitive to compression.}

\subsection{Multi-modal Information Fusion and Reasoning}


Suitable structure of knowledge representations allows \ac{genai} agents to compress, transfer, and retrieve knowledge efficiently and timely. 
However, information of the environment, other agents' states and actions obtained from observation space of different data modalities, e.g., images, audio, point could and \acf{rf} signals are generally correlated, thus introducing redundant information during transmission. 
Therefore, unifying the equivalent information from different data modalities or sources on a common semantic space is essential for efficient communication between \ac{genai} agents. 
For the sake of simplicity, we select natural language as the semantic space for information in which \acp{llm} effectively understand and manage. Meanwhile \acp{llm} can serve as a bridge to connect the knowledge and semantics representing the perceived information. Techniques such as contrastive learning (e.g., CLIP \cite{ramesh2022hierarchical}) and multi-modal cross-attention (e.g., ImageBind \cite{girdhar2023imagebind}) have been used to align multi-modal data, extendable to \ac{rf} signals as well. After training cross-modality encoders, semantics can be extracted from raw data uniformly. 


With knowledge and information represented in corresponding semantic space, an agent can perform planning and reasoning to make decisions until the ultimate goal or target is achieved. CoT prompting could be used to decompose a unseen tasks into coherent sequences serving as intermediate (known/solvable) steps, with an interpretable window suggesting how it might reach the final solution, and gives the agent opportunity to debug when a step goes wrong \cite{wei2022chain}. CoT could be generalized to sequential or non-sequential complex structures such as tree or graph of thoughts \cite{yao2024tree,besta2024graph}, with rewards to improve the reasoning path. Furthermore, the decomposed sub-tasks of a complex task can be assigned to different agents and be solved separately. Finally the outputs from each sub-task can guide agents to complete the task jointly. 

\begin{figure*}[t!]
\centering
\includegraphics[width=1\linewidth]{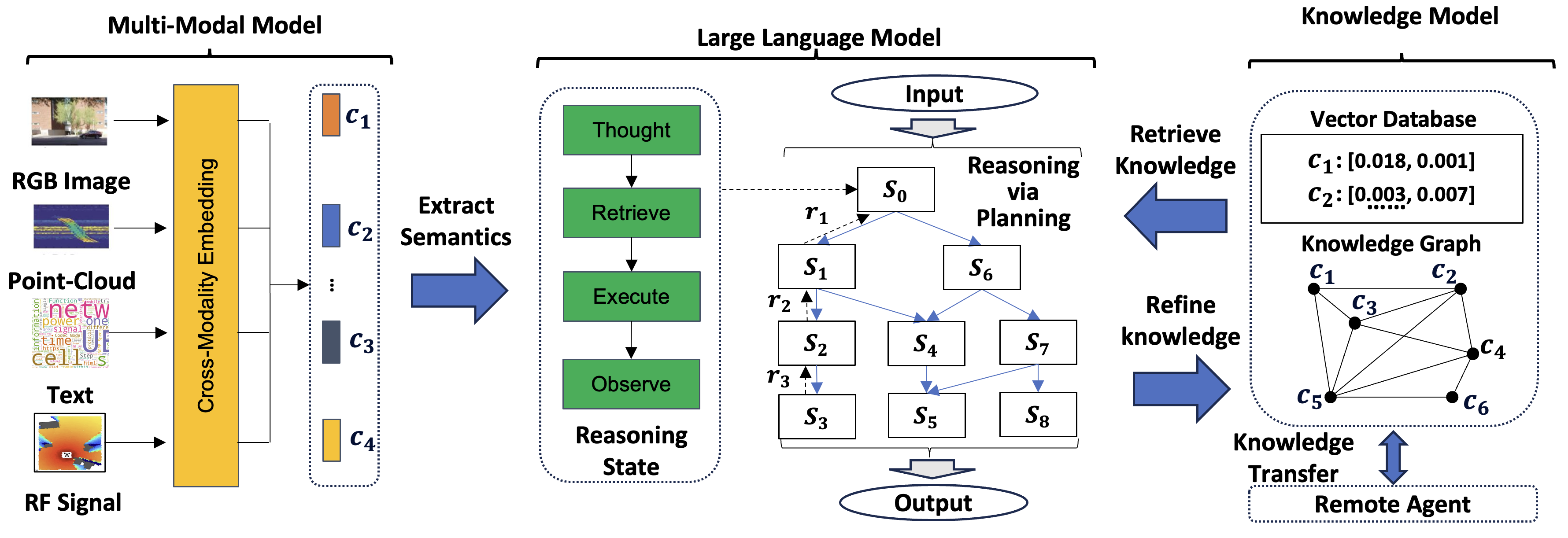}
\caption{
\textcolor{black}{Proposed pipeline of multi-modal semantic extraction, retrieval, and reasoning on GenAINet agents. Multi-modal LLMs unify information from various data modalities as semantics, enabling effective sensor data merging. Common and evolving knowledge are semantically represented (text or embeddings) to support compression, exchange, and retrieval, avoiding redundant multi-modal raw data exchange. LLMs reason and plan based on information, knowledge, and prior experiences, refining knowledge models over time. Our framework is highly adaptable to diverse tasks and heterogeneous data modalities.}}

\label{fig:multimodal_generation}
\end{figure*}

Fig. \ref{fig:multimodal_generation} describes the process of multi-modal semantic reasoning. Semantics are obtained from raw data by multi-modal \acp{llm} in a unified form. 
A knowledge model, where the semantic relations between knowledge are well defined in corresponding representation formats, is maintained and  refined through knowledge exchange from other agents if needed. When an agent needs to execute a task passively or proactively, it plans a tree of thought states based on the semantics of the perceived information and relevant knowledge extracted from the knowledge model jointly. In each state it retrieves relevant knowledge from the knowledge model, analyze the semantics provided by the multi-model model and then produce an reasonable action. The final output is generated after completing all planned states. The agent can observe reward at each state, improve the successive reasoning path and update knowledge in the knowledge if necessary. If one reasoning path yield good performance after the  full execution, \acp{llm} can analyze, summarize and then save it into the knowledge model as new evolving knowledge.

\subsection{Semantic-native GenAINet}

GenAI agents can collectively learn communication protocols and decision strategies for solving different tasks, leading to an efficient GenAINet. \ac{genai} agents can be used for various tasks on connected devices, from generating content to making decisions. An effective communication protocol should be part of the reasoning process, guiding agents to achieve a goal, such as the accuracy of responses, the accumulated reward of actions, and so on. Furthermore, an agent should effectively manage the knowledge to transmit, memorize, and update its model according to the information freshness and energy cost. 

There are two typical paradigms, namely, teacher-student or a distributed paradigm for GenAINet that can be implemented. The \emph{teacher-student paradigm}  is applicable for communication between cloud, edge, and device agents, 
where \emph{there is a dominant agent with considerably stronger capabilities compared to others}.
Specifically, a teacher agent with \ac{llm} trained on universal knowledge guides a student agent to perform specific tasks via knowledge transfer. To reduce communication cost, the teacher agent extracts and compresses knowledge with semantic equivalence in specific domains and transfer to the student, which is efficient for devices performing specific tasks (e.g., routing, traffic control). \textcolor{black}{In the centralized setting, smart grids represent an excellent use case for the teacher-student paradigm in GenAINet. In such a scenario, a teacher agent, deployed at a cloud or regional control center, is responsible for managing a large-scale electricity distribution network. This agent, powered by a highly capable \ac{llm}, continuously processes data from multiple edge nodes (e.g., smart meters, substations, renewable energy sources) to perform real-time load balancing, fault detection, and predictive maintenance. To enhance efficiency, the teacher agent extracts and compresses relevant semantic knowledge, such as energy consumption patterns, fault signatures, or renewable energy forecasts. This compressed knowledge is then transmitted to student agents located at edge nodes. The student agents leverage this tailored knowledge to execute localized tasks, such as regulating energy distribution in microgrids or predicting and responding to demand spikes in specific neighborhoods.} The distributed paradigm can be used for communication between mobile devices, machines, vehicles, where agents have similar capabilities and need to collaboratively complete tasks. Each agent has a memory of emergent evolving knowledge which is learned from its past experience. During planning, the agent can retrieve or communicate knowledge with others to optimize decisions and update local knowledge. In a long term, the agents will share a common knowledge model and make decisions locally, where communication costs can be minimized until observing new scenarios or tasks. Compared to standard communication systems, GenAINet is resource efficient and task effective. For example, in a remote query scenario the teacher-student paradigm can reduce load and latency in sending responses; in an \acf{v2x} scenario the distributed paradigm can improve driving safety and communication reliability. 

\begin{figure*}[t!]
\centering
\includegraphics[width=1\linewidth]{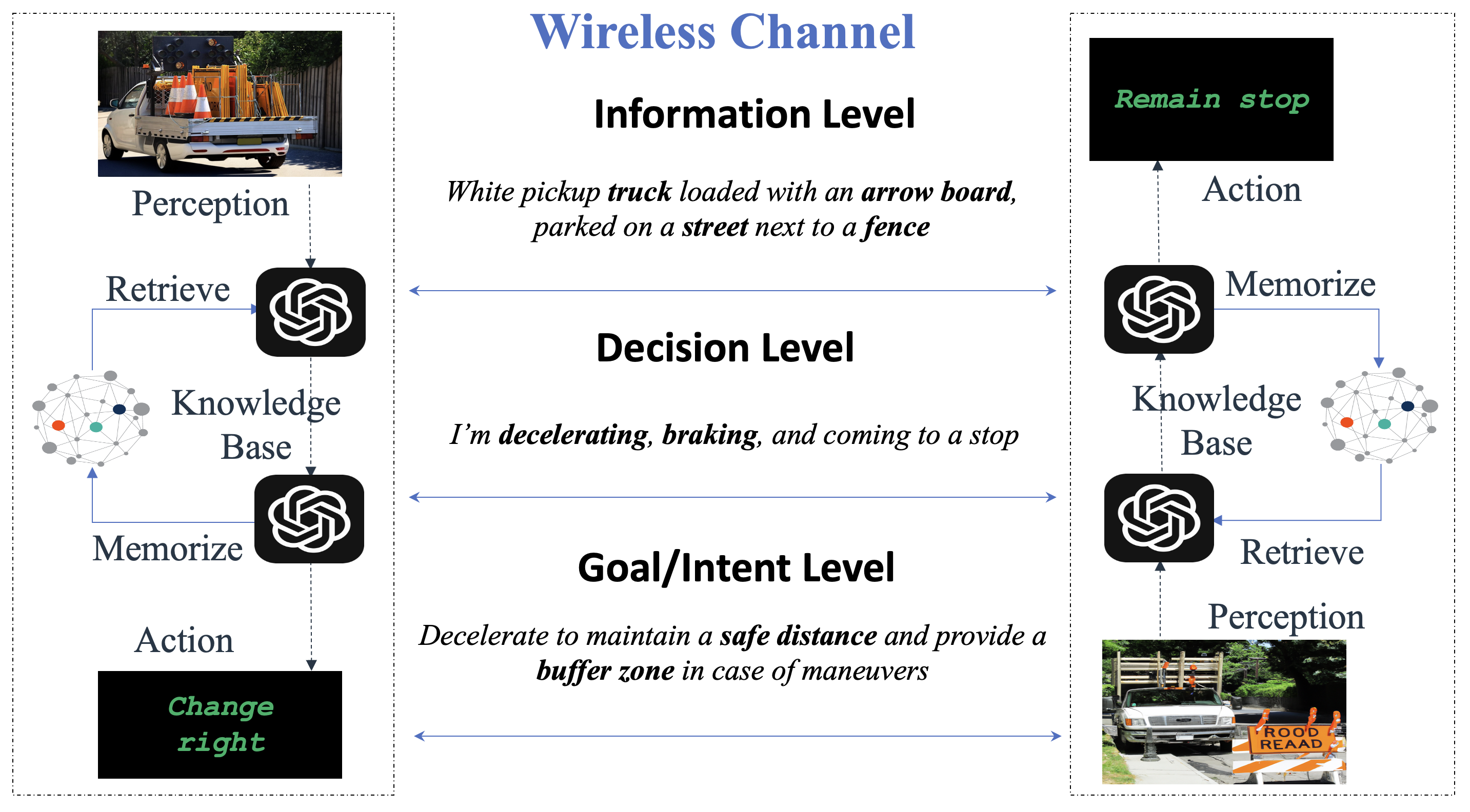}
\caption{
\textcolor{black}{Three levels of collaboration in GenAINet: {information}, {decision}, and {goal/intent} levels. At the information level, only relevant information messages are exchanged. At the decision level, agents share decisions while considering the decisions of others. At the goal/intent level, reasoning paths are generated by considering the goals of other agents, and only the goals are shared, reducing transmission overhead. For instance, car avoidance highlights how exchanging intent information rather than raw data like images suffices. These collaboration levels are adaptable to task complexity, reflecting human-like interactions in GenAINet.}}

\label{fig:collective_intelligent_vehicle}
\end{figure*}

Fig. \ref{fig:collective_intelligent_vehicle} illustrates an example of GenAINet communication between \acfp{av}. The car agent utilizes \acp{llm} as knowledge retriever (sender) and decision maker (actor). Communication is part of knowledge retrieval or sharing with a remote agent, which is planned according to actions and observations in a reasoning path. This makes the information exchange efficient and effective for decision making. The retriever aggregates information from local and remote knowledge model, pass it to the actor to generate driving commands. The actor optimizes reasoning path from environment feedback. For example, it can reward the \ac{cot} to find the best path and action. Once complete, the agent extracts knowledge (e.g., learned rules) from the recent planning and store it in the memory. 


Finally, we distinguish 3 different collaboration levels that can be achieved by \ac{genai} agents by combining the advantage of semantic communication and \acp{llm} reasoning. The first level is the {information level} from multi-modal sensory data. \acp{llm} can produce semantic latent on source textual data for lossless generation \cite{gilbert2023semantic}. With multi-modal \acp{llm}, semantic compression can be used to gather information efficiently from remote sensor with different field-of-views or resolutions, to improve downstream tasks such as accident detection, navigation. The second level is the {decision level}, where agents exchange and make their decisions jointly. 
This can be achieved by multi-agent debate framework, e.g., in \cite{liang2023encouraging,du2023improving}.
The third level is the {intent/goal level}, 
where the intent/goal of each agent is shared with other agents so that the common objective of all involved agents can be achieved jointly. For example, if the knowledge model of two agents are well synchronized, one agent merely needs to inform the another agent its intent/goal. Then the another agent can predict the decision better (even the whole plan of the former to a certain extent), leading to a huge reduction of communication costs.

\section{Case Studies of GenAINet} 

In this section, we show how the GenAINet can be applied to two examples namely, wireless device query and wireless power control. Specifically, we want to show: 1) How GenAI agents can compress semantically common knowledge in telecom context and transfer it to a wireless device to assist telecom domain knowledge query; 2) How GenAI agents can collaborate by sharing reasoning histories to solve a wireless network problem. 

\subsection{Semantic knowledge transfer for Telecom domain query}

\begin{figure*}[t!]
\centering
\includegraphics[width=1.0\linewidth]{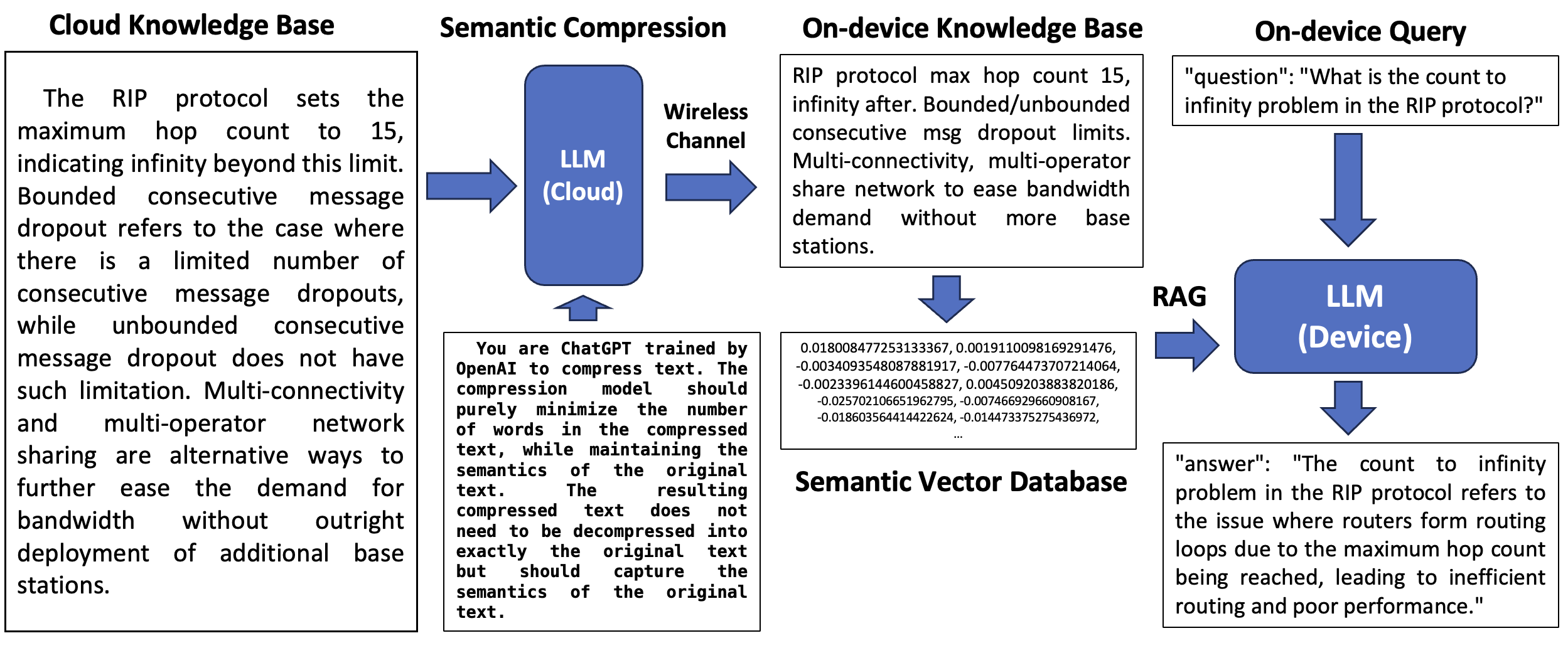}
\caption{Common knowledge transfer from a cloud \ac{llm} to an on-device \ac{llm} with semantic compression and \acfp{rag}. When an on-device \ac{llm} receives a knowledge query beyond its capabilities, it first searches for its semantic vector database to find the corresponding knowledge item with maximum semantic similarity and leverage \acp{rag} to generate accurate response. Notice that the on-device knowledge base is obtained by semantic compression of the huge cloud knowledge base.   }
\label{fig:remote_query}
\end{figure*}

Question answering (QnA) is a typical application of \acp{llm}. Deploying \acp{llm} close to the end user can largely reduce the latency and traffic burden from massive connections. Despite that light-weight \ac{llm}s are built with efficient inference techniques, they exhibit poor performance in specialized domains compared to regular \acp{llm}. \acp{rag} can enhance \ac{llm}'s knowledge with external database while introducing storage cost and higher latency. Envisioned from GenAINet, we propose a semantic knowledge driven on-device query. As shown in Fig. \ref{fig:remote_query}, in the cloud agent, its \ac{llm} extracts  the related context for the queries from cloud data and compress them into embedding vectors to build a semantic \ac{kb}. It is then sent to the \ac{llm} on a device to answer questions using RAG. The device can feedback questions to improve semantic extraction. 

We have performed experiments on the TeleQnA \cite{maatouk2023teleqna}, including 10k Telecom domain QnA from research and standard materials. We extract the questions' sources and build a large vector database in the cloud. An initial batch of sampled questions are fed into LLM to perform retrieval from the vector database, under instruction prompts to generate contexts related to the questions, and build the semantic KB for \acp{rag} on device. 

\textcolor{black}{Table \ref{tab:comparison} shows that the semantic KB significantly improves the TeleQnA accuracy in all categories compared to both full remote knowledge and local LLMs without external knowledge. In our experiments, the proposed semantic compression scheme achieves an average compression ratio of $45\%$, significantly reducing the amount of knowledge transferred from the cloud to local devices. Despite this compression, the accuracy drop is minimal—only $5\%$—compared to querying with full knowledge, while achieving a $14\%$ improvement in accuracy compared to local LLMs without external knowledge. These results highlight the effectiveness of the semantic compression approach in balancing communication efficiency and query accuracy. Notably, the trade-off between compression ratio and accuracy varies across QnA categories, reflecting differing sensitivities of knowledge domains to compression. For instance, lexicons maintain a high accuracy of $96\%$ even with nearly $50\%$ compression, whereas research publications experience a more significant accuracy drop under similar compression levels. Queries related to standards are comparatively less sensitive to semantic compression, with only a $3\%$ accuracy drop, while lexicons and research publications exhibit larger drops of $8\%$. These observations underscore the adaptability and versatility of the proposed framework, enabling significant communication savings while maintaining competitive performance across diverse query types. This variability suggests that compression ratios can be dynamically adjusted based on the query category, paving the way for an adaptive semantic compression scheme optimized with LLMs.}

\begin{table*}[t!]
\centering
\resizebox{0.9\textwidth}{!}{%
\begin{tabular}{c|c|c|c|c}
    \hline 
    \textbf{QnA Category} & \makecell{\textbf{Full Knowledge} (\%)} & \makecell{\textbf{Compressed Knowledge} (\%)} & \makecell{\textbf{No Knowledge} (\%)} & \makecell{\textbf{Compression Ratio} (\%)} \\
    \hline 
    Lexicon               & 92.00  & 96.00  & 88.00  & 48.03 \\
    Research overview     & 88.46  & 83.65  & 72.12  & 44.84 \\
    Research publication  & 91.16  & 83.25  & 71.63  & 40.05 \\
    Standard overview     & 96.77  & 91.94  & 77.42  & 47.65 \\
    Standard specification& 81.91  & 77.66  & 53.19  & 52.35 \\
    \hline
    \textbf{Overall}      & 89.60  & 84.00  & 69.20  & 45.52 \\
    \hline 
\end{tabular}%
}
\caption{\textcolor{black}{Accuracy and compression ratio (\%) of LLM agents for remote query with semantic compression, compared to query with full knowledge (upper bound) and query with LLM only (lower bound).}}
\label{tab:comparison}
\end{table*}

\subsection{Collaborative reasoning for wireless power control}

\textcolor{black}{Whereas the previous case study concerns telecom language capabilities of LLM agents, the present case study concerns their ability to tune a quantity in the radio domain namely, the transmit power of a wireless device. We consider power control here because it is a very famous problem, still of practical interest whenever formulated for new settings, makes easy the task of illustrating the concepts under consideration, and is not a priori a trivial task for an LLM agent since optimizing real numbers is involved. The conventional power control paradigm consists in modeling the environment, and then 
 finding (through optimization) the best power level leading to an assigned and fixed performance metric (data rate, energy, QoS, confidentiality, etc). With the proposed paradigm, the modeling and solving tasks are performed by the LLMs agents themselves (as performed in \cite{mongaillard-wiopt-2024} for EV charging policies). This is performed by exploiting the vast knowledge of the \acp{llm} about wireless networks (acquired during pre-training), to reason over the path from observed rate to best power \textcolor{black}{and cooperate with other agents through semantics, e.g., in natural language}. This capability in terms of flexibility is unique and has no equivalent in the vast literature of power control. This means in particular that with GenAI agents we may release online training and a wide range of performance metrics can be accommodated, depending on the wireless scenario or current network goals.}


We consider a scenario of \textcolor{black}{$N$} transmitter-receiver (Tx-Rx) pairs randomly placed in a $100 m^2$ squared area. Transmitters share a common spectrum where interference occurs. \textcolor{black}{The simulation assumes the absence of a central base station, reflecting scenarios in IoT settings where direct device-to-device (D2D) communication is preferred to reduce infrastructure costs and latency. Such decentralized architectures are particularly suitable for applications like smart grids or V2X networks, where devices operate autonomously and collaboratively without relying on a central controller. For simplicity, the channel gains $h_{ij}$ between every transmitter-receiver pair $(i,j)$ are taken to be realizations of Rayleigh law with a mean of $ \mathbb{E}(|h_{ij}|) = 1.0$. Each transmitter is initially assigned a transmit power $p_i$ randomly sampled from a uniform distribution over $[1,5]$ (in Watt). The system imposes a maximum power limit of $P_{\max}=10$ W. The communications are assumed to interfere each other, which translates into the signal-to-interference-plus-noise ratio (SINR) expression for Transmitter $i$: $\mathrm{SINR}_i = \frac{|h_{ij}|^2 p_i}{1+ \sum_{j\neq i}  |h_{ij}|^2 p_j }$. To reflect realistic transmission requirements, the target data rate for each transmitter is set as a scaled version of its achievable rate $\mu_i \log_2(1+ \mathrm{SINR}_i )$, with $\mu_i$ being a random scaling factor uniformly distributed over $[0.5,1.5]$. Transmitters have up to 10 rounds/iterations to adjust their power levels. The chosen reference Distributed Power Control (DPC) algorithm comes from \cite{foschini1993tvt}. Since the simulation setting is randomly generated, the feasibility conditions for DPC are not always satisfied, leading to complex and challenging scenarios for analysis. In the subsequent discussion, we focus on scenarios where the DPC algorithm fails to converge, highlighting the limitations and potential improvements for distributed power control using LLM-based solutions.} Furthermore, each Tx-Rx pair is associated with an \ac{llm} agent, including an \ac{llm} with plan-and-solve prompting, along with a memory of its past data rate observations and power allocation actions. Our goal is to reach a given data rate target while minimizing the sum-transmit power in the network. The radio environment information is unknown to the agents, and we instruct the \acp{llm} to exploit their memory and telecom knowledge. In each iteration, each \ac{llm} decides its power level and explains its decision.  \textcolor{black}{In our GenAINet framework, agents are empowered to propose, accept, or reject cooperation proposals expressed in natural language. The prompts used for individual GenAI agents and those incorporating cooperative mechanisms are shown in Fig. \ref{fig:power_template_reason} and Fig. \ref{fig:power_template_comm}, respectively.} 

\begin{figure}[t!]
\centering
\includegraphics[width=1.0\linewidth]{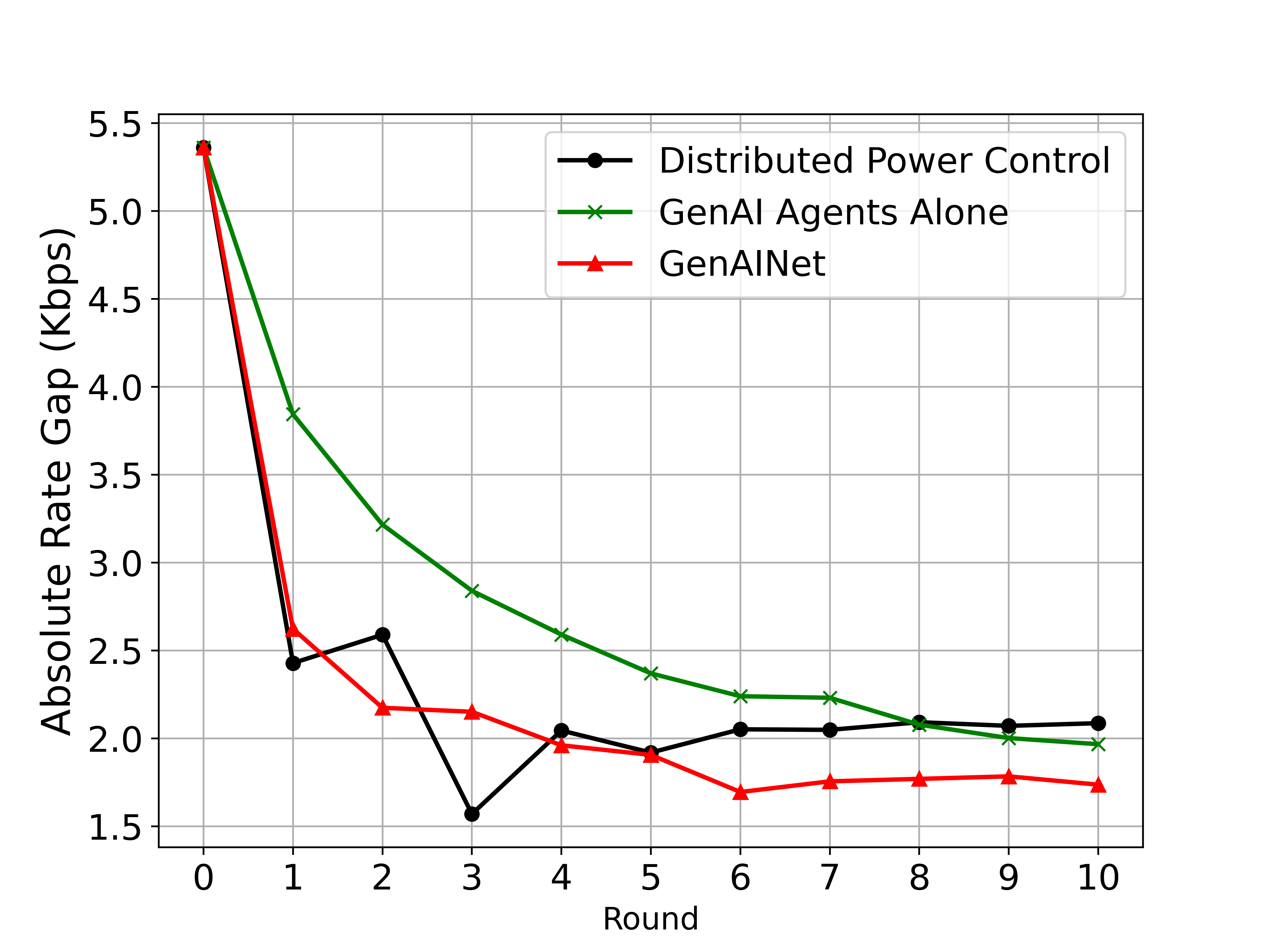}
\caption{\textcolor{black}{Absolute rate gap for the 4-user case versus the number of rounds for the distributed power control algorithm [34], the standalone approach where GenAI agents utilize memory and LLM reasoning, and the proposed GenAINet framework with natural language-based cooperation. Cooperation between GenAI agents results in faster convergence.}}
\label{fig:rate_gap_divergence}
\end{figure}

\begin{figure}[t!]
\centering
\includegraphics[width=1.0\linewidth]{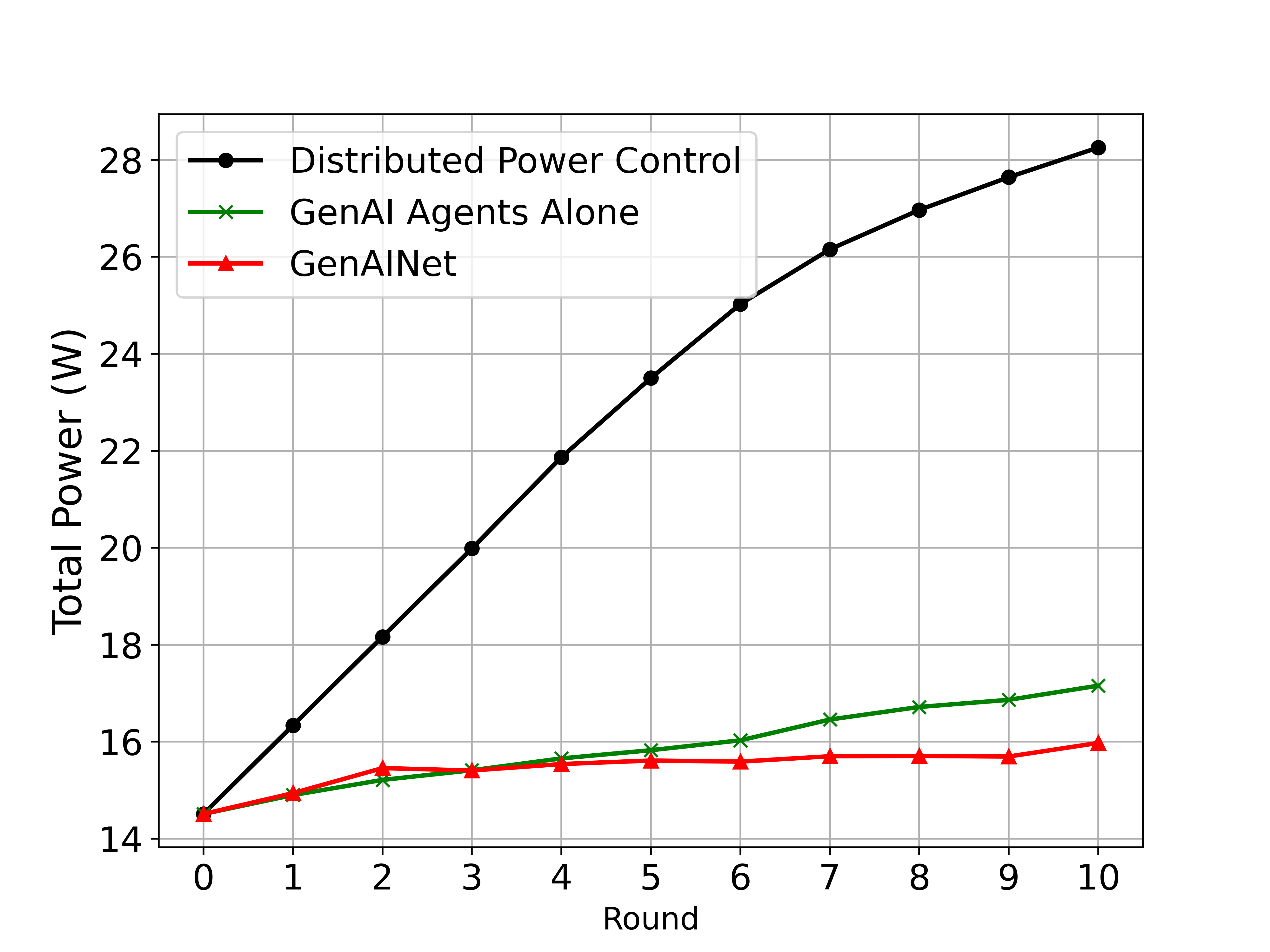}
\caption{\textcolor{black}{Total power for the 4-user case versus the number of rounds is presented for the distributed power control algorithm [34], the standalone approach where agents utilize memory and LLM reasoning, and the proposed GenAINet framework with natural language-based cooperation. Cooperation between GenAI agents leads to much smaller total power for preventing interference from growing rapidly.}}
\label{fig:total_power_divergence}
\end{figure}

\textcolor{black}{By defining the rate gap as the target rate minus the rate achieved after 10 rounds, Fig. \ref{fig:rate_gap_divergence} and Fig. \ref{fig:total_power_divergence} represent the rate gap and total power of the system for DPC, GenAI agents alone (with memory and reasoning capabilities), and the proposed GenAINet framework for $4$ Tx-Rx pairs. It can be observed that GenAI agents alone achieve a comparable rate gap to DPC without requiring online training, while maintaining significantly lower total power. Furthermore, when GenAI agents are allowed to cooperate through proposing and accepting natural language-based cooperation proposals, the GenAINet framework achieves a faster reduction in rate gap with even a slight decrease in power compared to GenAI agents alone. The increase in rate gap and total power for DPC arises from the selfish power adjustments of agents, which leads to a continual increase in interference. In contrast, GenAI agents with memory and reasoning tend to adjust their power levels more moderately to prevent drastic interference escalation. When cooperation is introduced, a subset of agents voluntarily maintain or slightly decrease their power levels, contributing to improved overall system performance.}

\textcolor{black}{We further compare the performance of different approaches for varying numbers of Tx-Rx pairs in Table \ref{tab:performance_comparison}. The proposed GenAINet approach consistently outperforms other approaches, except in the power level for the $2$-transmitter case. This observation suggests that cooperation may not be necessary when the number of transmitters is small. Additionally, the amount of exchanged information, measured as the number of messages sent per transmitter, is recorded in Table \ref{tab:performance_comparison}. On average, transmitters sent approximately $4$ to $5$ messages during the $10$-turn game, regardless of the number of Tx-Rx pairs, highlighting the scalability of the framework. It is worth noting that no prior knowledge of power control, online learning, or task-specific fine-tuning was required for the LLMs in our simulations. This example demonstrates the zero-shot adaptation and generalization capabilities of LLMs, effectively compensating for the potential high resource consumption of LLM-based solutions for complicated reasoning tasks.} 

\begin{figure}[t!]
\begin{mdframed}[linecolor=black, linewidth=2pt, roundcorner=10pt]
\textbf{System prompt:} Consider a wireless network with \{user\_num\} transmitter-receiver pair sharing the same spectrum. Each user has allocated bandwidth of \{bandwidth\} kHz. \textcolor{black}{Maximum power allowed is {Pmax}. The target rate of all Tx-Rx pairs: {rate} kbps.}

\vspace{0.1in}
\textbf{User prompt}: You are \{user\_id\}. Your current transmission power: \{power\} W. Your current transmission rate: \{rate\} kbps. You should adjust your transmit power to reach the targeted rate: \{target\} kbps \textcolor{black}{while help to minimize the total power of the network}. Think step by step, consider interference with others \textcolor{black}{and your past history}. Once you made a final decision, output in the following format: \{action:power, explanation:thought\}.

\vspace{0.1in}
\textbf{Memory}: \{user:user\_id, round:round\_num, observation:rate, action:power, explanation:thought\} 

\end{mdframed}
\caption{\textcolor{black}{Prompt template for collaborative power control leveraging GenAINet with memory and reasoning. The target of all agents is shared in the beginning then only each agent has accessed to its local memory and leverage it reasoning capabilities.}}
\label{fig:power_template_reason}
\end{figure}

\begin{figure}[t!]
\begin{mdframed}[linecolor=black, linewidth=2pt, roundcorner=10pt]
\textbf{System prompt:} Consider a wireless network with \{user\_num\} paired Tx and Rx users sharing the same spectrum. Each user has allocated bandwidth of \{bandwidth\} kHz. Maximum power allowed is {Pmax}. The target rate of all users: {rate} kbps.

\vspace{0.1in}
\textbf{User prompt}: You are \{user\_id\}. Your current transmission power: \{power\} W. Your current transmission rate: \{rate\} kbps. You should adjust your transmit power to reach the targeted rate: \{target\} kbps while help to minimize the total power of the network. Think step by step, consider interference and cooperation proposals with others. Be careful of not increasing drastically your power since it is damaging to increase the interference level for all Tx-Rx pairs. If you received proposals, your explanation should include the corresponding reasoning for accepting or rejecting them. You can propose cooperation plan to other users. If cooperation is needed, send a concise proposal to another user using this format: ``To User [id]: [content of proposal]".  Once you made a final decision, output in the following format: \{action:power, message:proposal explanation:thought\}.

\vspace{0.1in}
\textbf{Memory}: \{user:user\_id, round:round\_num, observation:rate, action:power, message:proposal, explanation:thought\} 

\end{mdframed}
\caption{\textcolor{black}{Prompt template for collaborative power control leveraging GenAINet with communication. The target of all agents is shared in the beginning, then each agent has accessed to its local memory, leverages it reasoning capabilities and sent/receive cooperation proposal. Agents are allowed to accept or rejected received proposals, leading to the a flexible and adaptive framework.}}
\label{fig:power_template_comm}
\end{figure}


\section{Challenges and Opportunities}

\begin{table*}[t!]
\centering
\resizebox{0.92\textwidth}{!}{%
\begin{tabular}{|c|c|c|c|c|c|c|c|}
    \hline
    \makecell{\textbf{Number of Transmitters}} & \multicolumn{3}{c|}{\textbf{Absolute Rate Gap (kbps)}} & \multicolumn{3}{c|}{\textbf{Total Power (W)}} & \textbf{Messages/Txs (GenAINet)} \\
    \cline{2-7}
     & \textbf{DPC} & \textbf{GenAI Alone} & \textbf{GenAINet} & \textbf{DPC} & \textbf{GenAI Alone} & \textbf{GenAINet} & \\
    \hline
    2 & 1.747 & 1.625 & 1.397 & 16.536 & 8.842 & 10.700 & 3.72 \\
    \hline
    4 & 2.086 & 1.967 & 1.736 & 28.254 & 17.156 & 15.974 & 5.00 \\
    \hline
    10 & 2.433 & 2.403 & 2.249 & 76.785 & 34.696 & 30.622 & 3.80 \\
    \hline
\end{tabular}%
}
\caption{\textcolor{black}{Performance comparison in terms of absolute rate gap, total power, and messages per transmitter (GenAINet only) for DPC, GenAI agents alone with memory and reasoning, and GenAINet allowing cooperation through natural language between agents.}}
\label{tab:performance_comparison}
\end{table*}

We envision GenAINet as a vital part for enabling \ac{ci} in 6G. However, there are several challenges to integrate \acp{llm} in wireless networks which opens up new research opportunities. 

\textbf{Edge \acp{llm}} The vast majority of existing \acp{llm} are deployed in the cloud, limiting their applicability on devices due to constraints pertinent to bandwidth, latency, and security. Despite their promising capabilities, deploying mainstream \acp{llm} on edge devices is extremely challenging. First, deploying large-scale \acp{llm} would be expensive for ordinary devices. 
Additionally, \acp{llm} have long inference time when solving sophisticated tasks, which significantly increases the communication latency. Finally, standards and protocols of wireless networks renew consistently, leading knowledge synchronization and updating of \acp{llm} energy-consuming through conventional techniques. Therefore it is crucial to develop efficient techniques to adapt \acp{llm} to wireless environment. Model compression, inference acceleration and selective knowledge sharing will mitigate these challenge potentially. \textcolor{black}{Model compression in particular allows significant reductions in terms of required memory to be achieved. Model compression relies on key techniques such as quantization \cite{jacob2018quantization}, pruning \cite{han2015deep}, distillation \cite{hinton2015distilling}, low-rank factorization \cite{hu2021lora}, parameter sharing \cite{ullrich2017soft}, weight clustering \cite{gong2014compressing}, sparse attention mechanisms \cite{beltagy2020longformer}, or dynamic inference techniques \cite{graves2016adaptive}. For instance, for quantization, significant gains can be obtained by quantizing float or double numbers (32 and 64 bits respectively) which are used to describe the LLM parameters. In the BitNet paper \cite{wang2023bitnet}, it is shown to quantize parameters with only 1.58 bit and obtain an energy consumption which is decreased by a factor of 71.4x. Also, efficient attention mechanism optimization techniques such as FlashAttention \cite{dao2022flashattention} and efficient inference techniques including PageAttention \cite{kwon2023efficient}, KV caching and prefill-decode separation will lead to a considerable reduction of inference time. Finally task/goal-oriented compression techniques helps to further reduce the communication overhead between agents \cite{zou2022goal}.} 

\textbf{Efficient Adaptation} \acp{llm} are auto-regressive generative models pre-trained on natural text to predict the next token. Therefore \acp{llm} hardly predict high-level semantics, which introduces high information redundancy and computation cost in training and inference. Besides, \acp{llm}' generalization capability is usually limited in domains not included in the training data. RAG and fine-tuning are commonly used to enhance \acp{llm}' knowledge while being expensive and inflexible. Furthermore, grounding \acp{llm}' knowledge to real world representation is challenging. Reinforcement learning is studied to optimize LLM online, which is not suitable for computational limited devices.

 \textbf{World Model} is a hierarchical, modular model predicting future representations of the state of the world \cite{lecun2022path}. It is trained to predict the high-level abstract instead of raw data. For example, I-JEPA \cite{assran2023selfsupervised} is trained to predict missing embedding on an image, which shows less training data and computational effort. Furthermore, hierarchical JEPA (H-JEPA) allows to learn higher level abstract representation which is effective for long-term prediction and eliminate the irrelevant details. Built on H-JEPA, hierarchical planning can be done by predicting state transition on abstract space, which could handle uncertain environment with minimal cost. JEPA framework gives promises for deploying world model on wireless devices than \acp{llm}. Since RF propagation is more abstract than text, training RF-JEPA could be more efficient than RF-GPT.  However, JEPA on multi-modality and planning to enable \ac{ci} still require further research.

 \textbf{\ac{rf} Foundation Models} While GenAI agents are used in various domains, applying them in wireless networks is still challenging. First, a network has a complex hierarchical architecture from RF to service layers, making LLMs difficult to decompose tasks for every network element. Second, orchestrating numerous agents' behavior is difficult for networks deployed in large scale geographical areas. Therefore, current \acp{llm} are used only in small domains, mostly service layer. Moreover, future networks require much higher reliability and robustness, where the uncertainty of \acp{llm} is problematic. Finally embedding RF signals into \acp{llm} is challenging, primarily due to the unavailability of appropriate large datasets and the inherent nature of RF signals, which is both spectral and spatial. This fundamentally differs from the textual data.

\section{Conclusion}
In this paper, we introduced the GenAINet framework, a knowledge-driven communication and reasoning network, as a promising enabler for collective intelligence. The architecture of GenAINet leverages unified \ac{llm}-powered GenAI agents to optimize both network protocols and applications. Agents in our semantic-native framework can extract unified semantics from \textcolor{black}{heterogeneous} raw data in the environment or from other agents, and leverage a semantic knowledge model to achieve effective communication and reasoning. Agents utilize \acp{llm} with semantics retrieved locally and remotely through semantic communications to effectively plan reasoning paths and make informed decisions for a task. We investigated a use case involving the compression and transfer of common knowledge between two agents to improve on-device query performance with reduced communication costs, as well as an example where distributed agents communicate using reasoning to \textcolor{black}{accomplish general} wireless power control \textcolor{black}{problems without predefined communication protocols}. We demonstrated that our multi-agent GenAINet can unleash the power of collective intelligence \textcolor{black}{with flexible levels of collaboration, much like human interactions}, and highlighted some fundamental challenges and future directions for leveraging \acp{llm} in 6G networks to enable collective intelligence. \textcolor{black}{Of course, challenges remain, such as making the proposed framework scalable. In particular, having sufficiently small and fast LLMs is necessary for their massive deployment in wireless networks. Moreover, as both a fundamental and practical challenge, it is imperative to design LLMs that allow some guarantees in terms of QoS and efficiently manage energy consumption.}

\bibliographystyle{IEEEtran}
\bibliography{ref}

\begin{IEEEbiography}[{\includegraphics[width=1in,height=1.25in,clip,keepaspectratio]{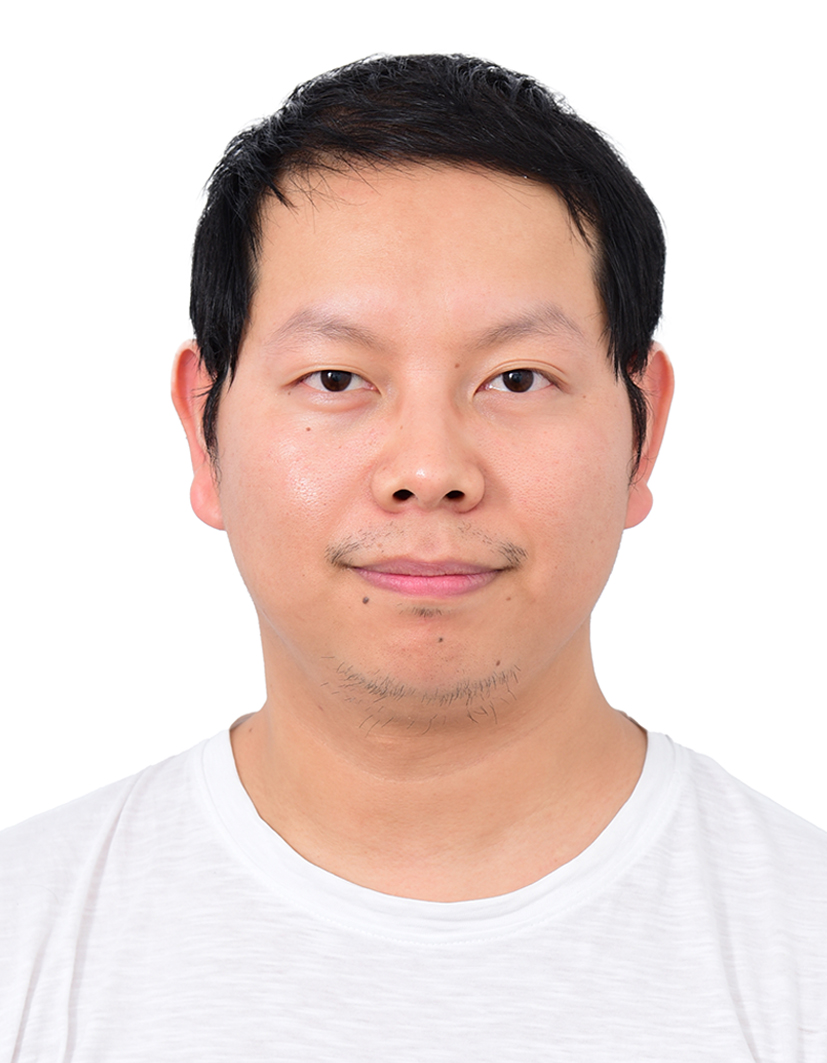}}]{Hang Zou} Hang Zou received the M.Sc. and Ph.D. degrees in Wireless Communications from Paris-Saclay University, France, in 2018 and 2022, respectively. Since October 2022, he has been with Technology Innovation Institute, UAE, where he is currently a Researcher. His main research interests include semantic communications, goal-oriented communications, generative AI and large language models.
\end{IEEEbiography}

\begin{IEEEbiography}[{\includegraphics[width=1in,height=1.25in,clip,keepaspectratio]{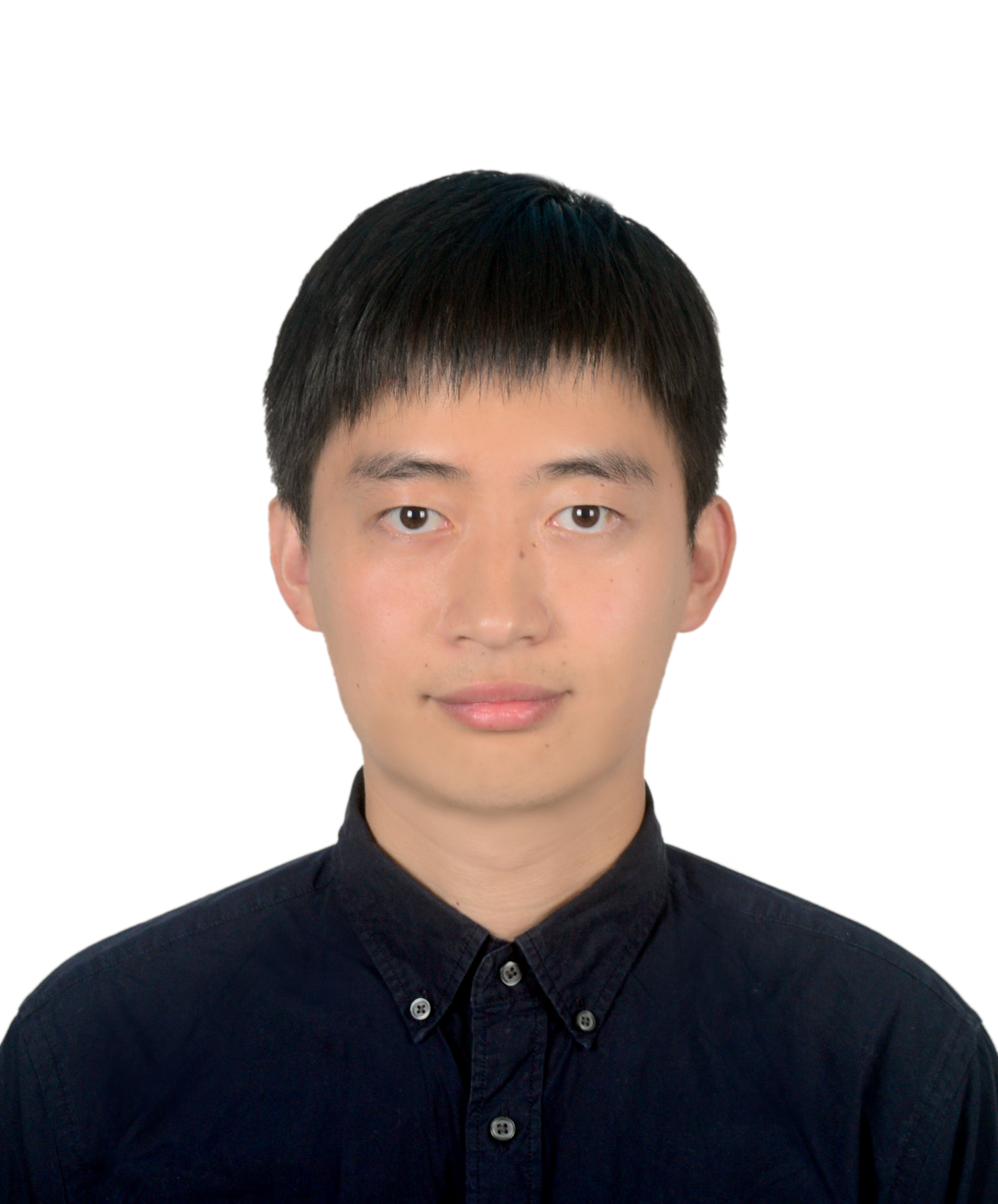}}]{Qiyang Zhao} Qiyang Zhao received B.S. degree in Information Security in 2009 from Xidian University, China and Ph.D. degree in Electronic Engineering in 2013 from University of York, UK, respectively. He has been working with industry on research, development, and standardization of 5G and 6G wireless communication systems. Since February 2022, he has been with Technology Innovation Institute, UAE, where he is currently a Lead Researcher. His research interests cover various aspects of artificial intelligence and telecommunications, with current emphasis on native AI network, large language model, semantic communication, multi-agent system, and multi-model learning.
\end{IEEEbiography}

\begin{IEEEbiography}[{\includegraphics[width=1in,height=1.25in,clip,keepaspectratio]{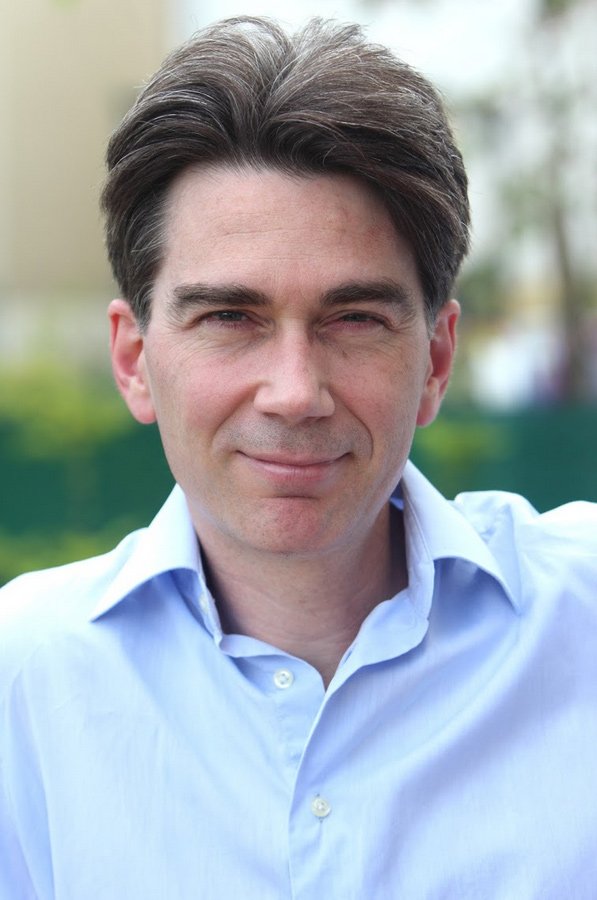}}]{Samson Lasaulce} (IEEE Member) Samson Lasaulce is currently a Chief Research Scientist in AI with Khalifa University (KU), Abu Dhabi. He is the holder of the TII 6G Chair on Native AI. He is also a CNRS Director of Research with CRAN at Nancy (joint lab between CNRS and University of Lorraine). He has been the holder of the RTE Chair on the Digital Transformation of Electricity Networks. He has also been a Professor with the Department of Physics at Ecole Polytechnique. Before joining CNRS he has been working for five years in private R$\&$D companies (Motorola Labs and Orange Labs). Dr. Lasaulce is the recipient of several awards. Dr. Lasaulce has been serving as an Associate Editor for several journals such as the IEEE Transactions on Signal Processing. His current research interests lie in large language models for optimization, transfer learning, model compression, distributed optimization, optimal control, and game theory. 
\end{IEEEbiography}

\begin{IEEEbiography}[{\includegraphics[width=1in,height=1.25in,clip,keepaspectratio]{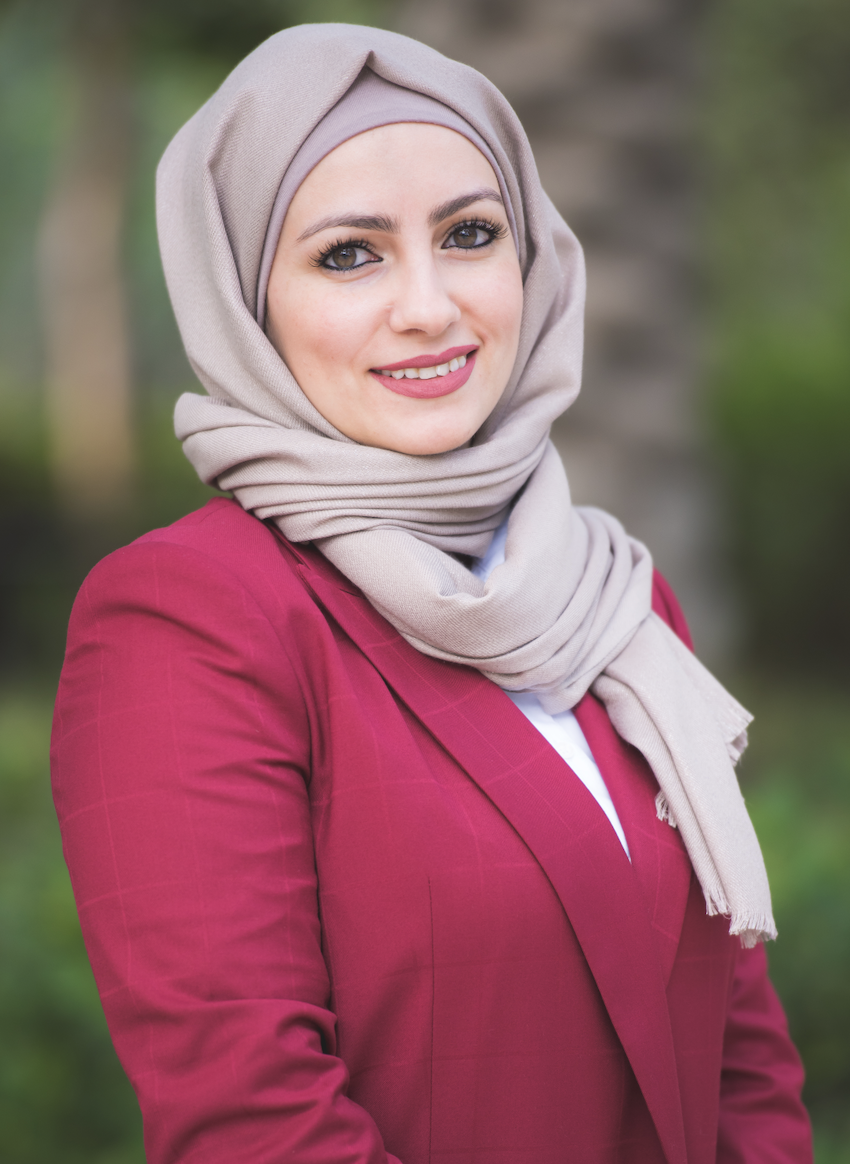}}]{Lina Bariah} Lina Bariah received the M.Sc. and Ph.D. degrees in communications engineering from Khalifa University, Abu Dhabi, UAE, in 2015 and 2018, respectively. She was a Visiting Researcher with the Department of Systems and Computer Engineering, Carleton University, Ottawa, ON, Canada, in 2019, and an affiliate research fellow, James Watt School of Engineering, University of Glasgow, UK. She was a Senior Researcher at the technology Innovation institute. She is currently an Adjunct Professor at Khalifa University, and an Adjunct Research Professor, Western University, Canada. Dr. Bariah is a senior member of the IEEE, IEEE Communications Society, IEEE Vehicular Technology Society, and IEEE Women in Engineering. She is the founder and lead of Women in Machine Learning and Data Science (WiMLDS)-Abu Dhabi Chapter. She was recently listed among the100 Brilliant and Inspiring Women in 6G", by Women in 6G organization. She is an Editor at IEEE Transactions on Wireless Communications. She was an Associate Editor for the IEEE Communication Letters, an Associate Editor for the IEEE Open Journal of the Communications Society, and an Area Editor for Physical Communication (Elsevier). She is a Guest Editor in IEEE Communication Magazine, IEEE Network Magazine, and IEEE Open Journal of Vehicular Technology.
\end{IEEEbiography}

\begin{IEEEbiography}[{\includegraphics[width=1in,height=1.25in,clip,keepaspectratio]{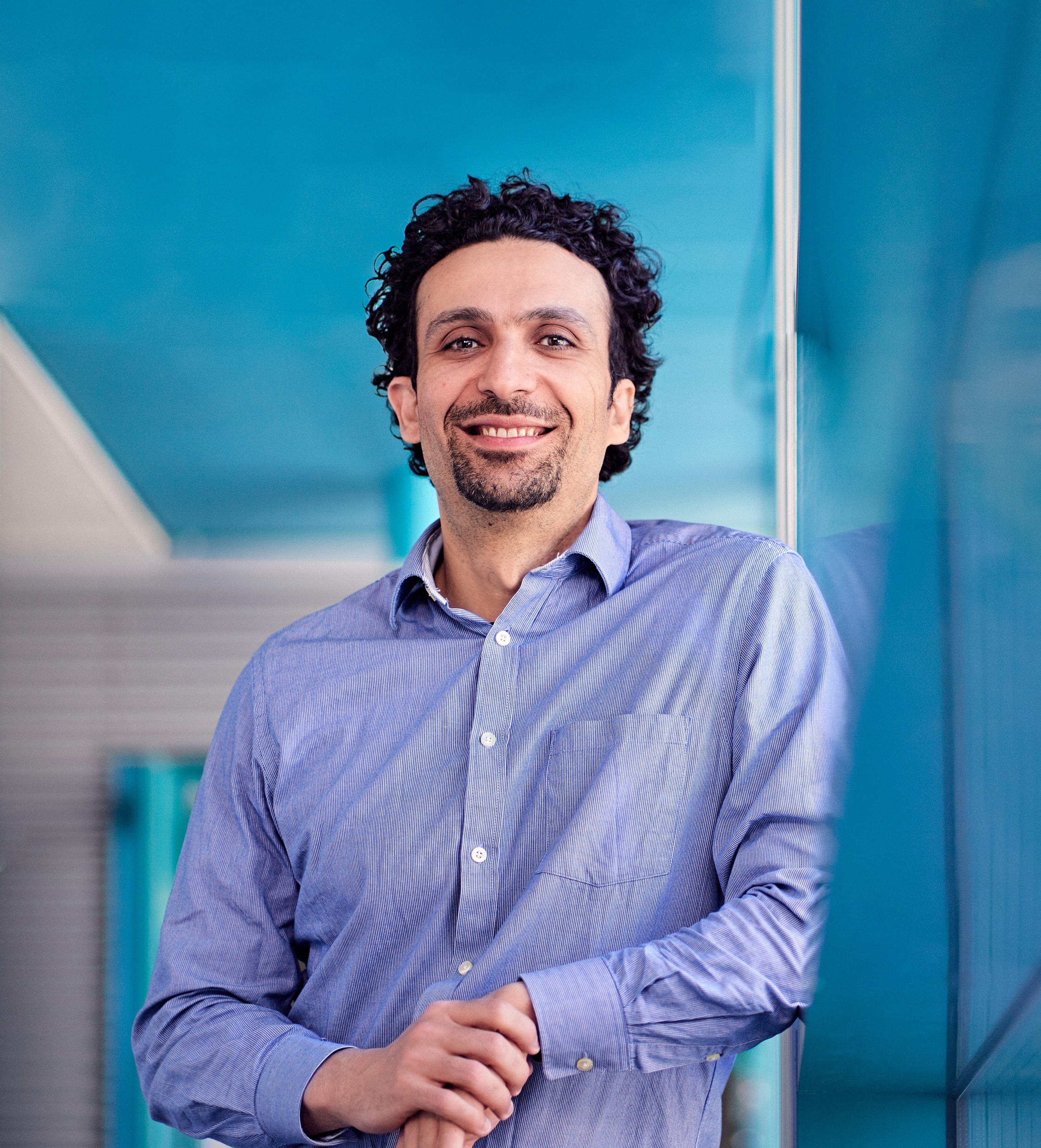}}]{Mehdi Bennis} (IEEE Fellow) Mehdi Bennis is a Professor at the Centre for Wireless Communications, University of Oulu, Finland, IEEE Fellow and head of the intelligent connectivity and networks/systems group (ICON). He has published more than 400 research papers in international conferences, journals and book chapters. He has been the recipient of several prestigious awards including the 2015 Fred W. Ellersick Prize from the IEEE Communications Society, the 2016 Best Tutorial Prize from the IEEE Communications Society, the 2017 EURASIP Best paper Award for the Journal of Wireless Communications and Networks, the all-University of Oulu award for research, the 2019 IEEE ComSoc Radio Communications Committee Early Achievement Award and the 2020-2023 Clarviate Highly Cited Researcher by the Web of Science.
\end{IEEEbiography}

\begin{IEEEbiography}[{\includegraphics[width=1in,height=1.25in,clip,keepaspectratio]{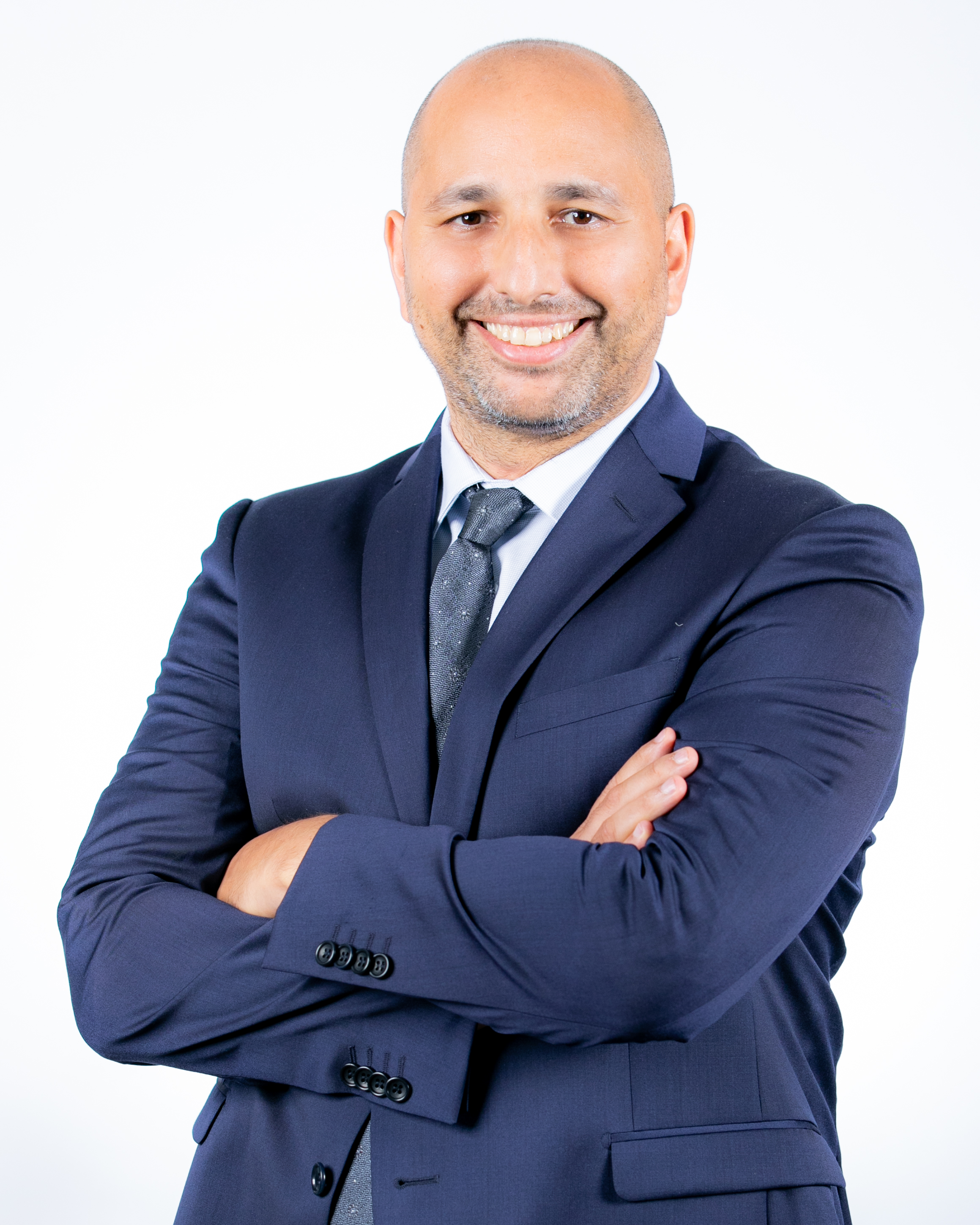}}]{Merouane Debbah} (IEEE Fellow) Mérouane Debbah is Professor at  Khalifa University of Science and Technology in Abu Dhabi and founding Director of the KU 6G Research Center. He is a frequent keynote speaker at international events in the field of telecommunication and AI. His research has been lying at the interface of fundamental mathematics, algorithms, statistics, information and communication sciences with a special focus on random matrix theory and learning algorithms. In the Communication field, he has been at the heart of the development of small cells (4G), Massive MIMO (5G) and Large Intelligent Surfaces (6G) technologies. In the AI field, he is known for his work on Large Language Models, distributed AI systems for networks and semantic communications. He received multiple prestigious distinctions, prizes and best paper awards (more than 40 IEEE best paper awards) for his contributions to both fields. He is an IEEE Fellow, a WWRF Fellow, a Eurasip Fellow, an AAIA Fellow, an Institut Louis Bachelier Fellow, an AIIA Fellow  and a Membre émérite SEE.
\end{IEEEbiography}

\EOD

\end{document}